%% file: main.tex
\documentclass[]{spie}  

\usepackage{url}      
\usepackage{float}
\usepackage{graphicx}
\usepackage{tabularx}
\usepackage{makecell}
\usepackage{array}
\usepackage{subcaption}
\usepackage{caption}
\usepackage{amsmath,amsfonts,amssymb}

\usepackage{graphicx}
\usepackage[colorlinks=true, allcolors=blue]{hyperref}

\title{Monocular Microscope to CT Registration using Pose Estimation of the Incus for Augmented Reality Cochlear Implant Surgery}

\author[a]{Yike Zhang}
\author[a]{Eduardo Davalos}
\author[a]{Dingjie Su}
\author[b]{Ange Lou}
\author[a,b]{Jack H. Noble}
\affil[a]{Dept. of Computer Science, Vanderbilt University}
\affil[b]{Dept. of Electrical and Computer Engineering, Vanderbilt University}

\begin{document}
\maketitle

\input{abstract}

\input{introduction}

\input{methods}

\input{results}

\input{conclusion}

\input{acknowledgement}

\bibliography{citation.bib}
\bibliographystyle{spiebib} 

\end{document}

%% file: abstract.tex
\begin {abstract}
For those experiencing severe-to-profound sensorineural hearing loss, the cochlear implant (CI) is the preferred treatment. Augmented reality (AR) aided surgery can potentially improve CI procedures and hearing outcomes. Typically, AR solutions for image-guided surgery rely on optical tracking systems to register pre-operative planning information to the display so that hidden anatomy or other important information can be overlayed and co-registered with the view of the surgical scene. In this paper, our goal is to develop a method that permits direct 2D-to-3D registration of the microscope video to the pre-operative Computed Tomography (CT) scan without the need for external tracking equipment. Our proposed solution involves using surface mapping of a portion of the incus in surgical recordings and determining the pose of this structure relative to the surgical microscope by performing pose estimation via the perspective-n-point (PnP) algorithm. This registration can then be applied to pre-operative segmentations of other anatomy-of-interest, as well as the planned electrode insertion trajectory to co-register this information for the AR display. Our results demonstrate the accuracy with an average rotation error of less than 25 degrees and a translation error of less than 2 mm, 3 mm, and 0.55\% for the x, y, and z axes, respectively. Our proposed method has the potential to be applicable and generalized to other surgical procedures while only needing a monocular microscope during intra-operation.

\end {abstract}

\keywords{image-guided surgery, pose estimation, surface mapping, intra-operative registration, deep learning}

%% file: introduction.tex
\section{INTRODUCTION}
\label{sec:intro}

Cochlear Implant (CI) uses an array of surgically implanted electrodes in the cochlea to directly stimulate the auditory nerve and induce the sensation of hearing. CI surgery involves a wide field mastoidectomy to gain access to the cochlea and avoid damaging nearby sensitive structures. CI outcomes are variable, and the intra-cochlear positioning of the electrode array is a key factor that is known to have an impact\cite{holden2013factors}. Previous studies have demonstrated that image guidance can lead to improved intra-cochlear positioning of CI electrode arrays \cite{labadie2018preliminary}.  In that approach, the electrode insertion trajectory is planned based on anatomical segmentation in a pre-implantation CT scan, and the plan is presented to the surgeon with a 3D graphical user interface and as a set of instructions regarding the position of the optimal insertion trajectory relative to adjacent anatomy that can be visually referenced in the Operation Room (OR). AR is another guidance approach that eliminates the need for the surgeon to memorize plan details or to have to avert attention from the surgical scene to consult the planning user interface on a separate computer. AR can be facilitated in the already used surgical microscopes for CI with the use of an integrated microdisplay that allows the injection of information directly into the surgeon’s optics via a beamsplitter to overlay information on the view of the surgical scene. Typically, AR solutions for image-guided surgery rely on optical tracking systems (e.g., BrainLab) to register pre-operative planning information on the microscope so that hidden anatomy or other information can be overlayed co-registered with the view of the surgical scene \cite{vavra2017}. 

Our goal is to develop a method that permits direct 2D-to-3D registration of the microscope video to the pre-operative CT scan without the need for external tracking equipment. This would reduce the cost of integrating AR solutions into the clinical workflow and has the potential to be more accurate, since the microscope has micron-level resolution, whereas optical trackers typically have target registration errors on the order of 0.5 mm. The proposed solution involves identifying and surface mapping a portion of the ossicles that is consistently visible both in CI surgery and CT scans, then determining the pose of this structure relative to the microscope by using PnP to achieve 2D-to-3D registration. This solution can be applied to pre-operative segmentation of other hidden anatomy, as well as the planned electrode insertion trajectory to co-register this information to the OR for AR display. We choose the head of the ossicles as the target structure because, despite limited visibility, it is a key structure we can identify that is consistently at least partially visible in the microscope and identifiable in pre-operative CT. Most of the remaining visible anatomy in the microscope has been surgically altered when creating the mastoidectomy and no longer has clear correspondence to the pre-operative CT. 

Locating the poses of objects within cluttered scenes is a critical task in many computer vision and robotics-related applications. 6D pose estimation, a field focused on determining the rotation matrix and translation vector parameters of an object in camera-centered coordinates, has long been a research area of interest, yielding noteworthy contributions over the years. The prevalent methods are based on deep learning and generally fall into two categories: instance-level object pose estimation \cite{dblp} and category-level object pose estimation \cite{nocs}. The presented methods typically rely on four different types of input: images from monocular, stereo, and depth cameras, as well as point clouds from LIDAR. These techniques have achieved impressive performance in household and outdoor settings. However, further discussion of these techniques applied in surgical scenarios remains relatively unexplored. This may be attributed to the unique challenges raised by surgical video datasets, including insufficient data, textureless objects, limited visibility of key structures, and the complex data acquisition process. All of these factors make the implementation of commonly used pose estimation techniques particularly challenging. Furthermore, it is generally acknowledged that estimation based solely on monocular images tends to yield less accuracy compared to stereo images, as the latter provides additional depth information. 

In this work, we focus on monocular images because they are currently more widely used in Otology. In this work, we address the limitations mentioned above and extend the pose estimation schemes of popular household scenarios to be more suitable and applicable to the medical domain. Our approach employs a lightweight, customized neural network, specifically tailored to scenarios where the landmark object is heavily occluded throughout the procedure in a cluttered scene. This paper presents our primary method, preliminary results, and potential implications for future applications in surgical procedures. We believe that our work represents a step forward towards addressing the issue of pose estimation in heavily occluded and cluttered surgical environments, potentially paving the way for AI-powered augmented reality in various types of surgery.

%% file: methods.tex
\section{METHODS}
\label{sec:method}
In this section, we present our data preparation process and a two-stage training pipeline, as shown in Fig. \ref{fig:two_stage}.
\input{figures/method_two_stage}
\noindent The primary task is to estimate the pose of the occluded ossicles within cluttered surgical scenes. Traditional keypoint-based registration strategies would be problematic in this particular scenario due to severe occlusion and limited visibility of our target. One common way to track occluded objects is to establish a 2D-3D surface mapping with visible landmark structures\cite{dblp}. Therefore, we propose to create this surface mapping of the ossicles head to create a coordinate map for our application. Our dataset includes twelve surgical videos and their corresponding CT scans with segmentation of the structures of the ear. To obtain a segmentation of the patient’s ossicles, which includes the incus, we apply a deep learning based registration method\cite{zhang2023self} to align the patient’s CT with the atlas CT scan, subsequently applying the resulting deformation field to morph the atlas ossicles surface mesh into patient-coordinates. Within these twelve meshes, we created a universal 2D coordinate map for the short process of the incus for this ossicles due to the point-to-point correspondence characteristic in these meshes. This map defines the latitude and longitude coordinates for each vertex in the short process of the incus and part of the mallus head. To create this, on the surface of the atlas mesh, we define a north pole vertex $\alpha$ and a south pole vertex $\beta$ and calculate the normalized latitude value $\mu$ for each other vertex based on their geodesic distance to $\alpha$ and $\beta$. 
A prime meridian $p$ was calculated, finding the shortest geodesic path between $\alpha$ and $\beta$. We then calculate the normalized longitude value $\nu$ for each vertex based on its geodesic distance to the left and right sides of $p$. The $\mu$ and $\nu$ are calculated using Eq. \ref{eq:geo}.
\input{equations/geo}

\input{figures/coordinates_map}
\noindent The coordinate map is defined for the short process of the incus and a small portion of the malleus head that is potentially visible throughout surgery. Given that all patient meshes were defined via non-rigid registration from the same atlas mesh, they inherently maintain point-to-point correspondence, allowing us to universally apply the same latitude-longitude map, as illustrated in Fig.\ref{fig:coordinate_map} with latitude in red and longitude in green. This makes our method consistent with a category-level approach, as it learns across multiple instances to robustly predict unseen ossicles\cite{nocs}.

We adapt a pre-trained neural network (e.g., medium YOLOv8\cite{Jocher_YOLO_by_Ultralytics_2023}) to detect a bounding box around and segment the visible portion of the incus from each frame in our video dataset. Fig. \ref{fig:video_frame} shows three samples that are randomly selected from our total of twelve cases. Figs. \ref{subfig:o1}, \ref{subfig:o2}, \ref{subfig:o3} show the original images. Figs. \ref{subfig:seg1}, \ref{subfig:seg2}, \ref{subfig:seg3} highlight an example of segmented ossicles with a bounding box processed by fine-tuned YOLOv8. Since the microscope is not tracked, to determine ground-truth microscope extrinsic matrices, we developed an open-source software package that we call “Vision6D,” to facilitate manual registration of 3D objects, such as the ossicles and other structures, to 2D projection images, thus creating the ground-truth labels\cite{Zhang_Vision6D}. Fig. \ref{subfig:register1}, \ref{subfig:register2}, \ref{subfig:register3} shows manually registered ossicles in their corresponding video frame with an estimated focal length equal to 50,000 millimeters. Other structures of the ear, including the facial nerve, chorda, and scala tympani, are colored gray. Figs. \ref{subfig:visible1}, \ref{subfig:visible2}, \ref{subfig:visible3} demonstrate the visible surface of the ossicles with coordinate mapping on the surgical scene. As can be seen in panels Fig. \ref{subfig:register1}, \ref{subfig:register2}, \ref{subfig:register3} and Fig. \ref{subfig:visible1}, \ref{subfig:visible2}, \ref{subfig:visible3}, only a small portion of the incus is visible and is not occluded by other structures.
\input{figures/video_frame} 
\noindent Given the minimal camera movement inherent in cochlear implant surgeries, we opted to train our neural network using a selected frame of each surgical video. Our training dataset contains nine video frames, while our validation dataset includes three, each from unique surgical cases. To enhance our training dataset size, we use data augmentation techniques such as flipping, rotating, and translating, thus expanding our data size by a factor of 1000 with synthetically generated data. We propose a UNet-based  neural network to predict the coordinate mapping of the segmented region of visible incus. The inputs to the network are patches of the intersection of the monocular microscope image and incus segmentation. The patches are scaled down to a size of 108 $\times$ 192 and cropped to reduce the input size to 64 $\times$ 64. This process inherently reduces the number of pixel values in the correspondence map that need to be regressed, thus decreasing the overall computational time. We propose three loss terms with equal weights used when training the network: BCE loss, MSE loss, and Structural Similarity Index Measure (SSIM) loss, shown below.
\input{equations/loss}

%% file: figures/method_two_stage.tex
\begin{figure}[H]
    \centering
    \includegraphics[width=.95\textwidth]{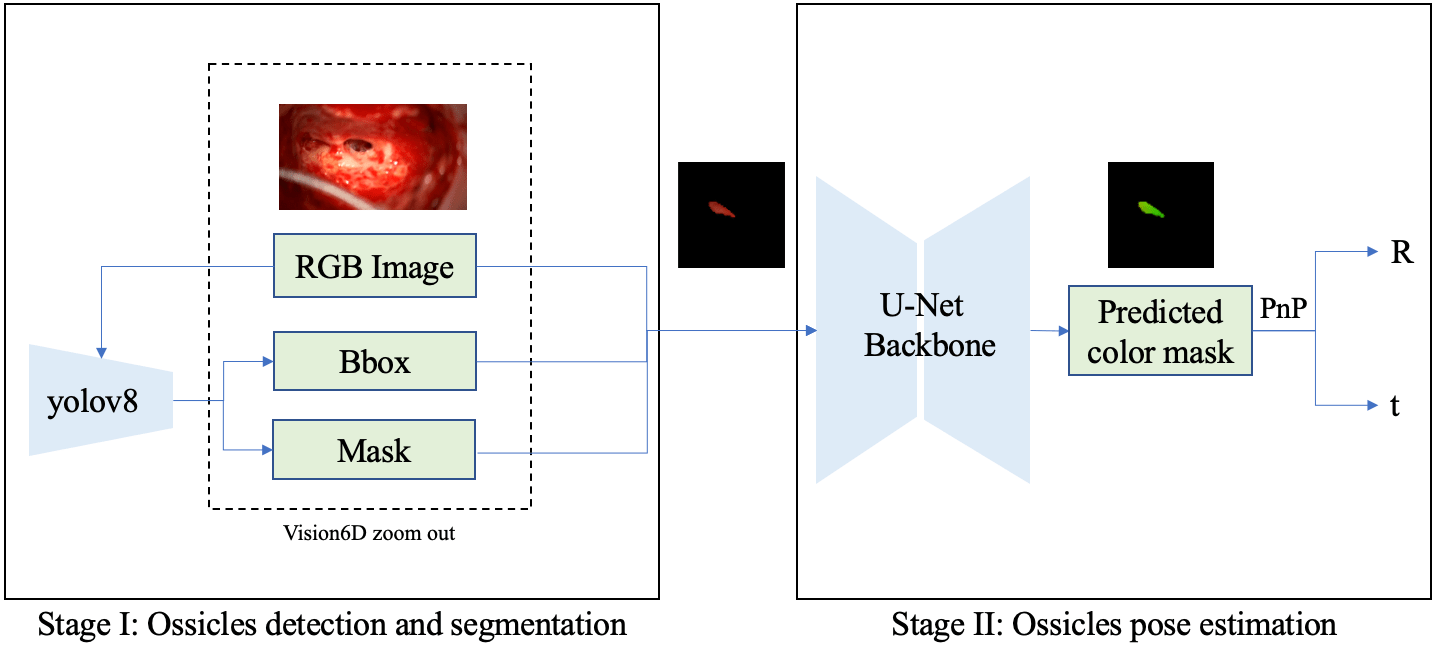}
    \caption{Two-stage Training pipeline}
    \label{fig:two_stage}
\end{figure}

%% file: equations/geo.tex
\begin{equation}
(\mu, \nu) = \left(\frac{d_\alpha}{d_\alpha + d_\beta}, \frac{d_l}{d_l + d_r}\right)
\label{eq:geo}
\end{equation} where $d_\alpha$ and $d_\beta$ are the geodesic distance from the vertex to the north and south poles, while $d_l$ and $d_r$ are the geodesic distance from the vertex to the left and right sides of the prime meridian $p$. They are computed using a fast marching-based \cite{Sethian} solution to the eikonal equation $|\nabla U| = 1$ on the ossicles surface.

%% file: figures/coordinates_map.tex
\begin{figure}[H]
    \begin{subfigure}[b]{0.3\textwidth}
        \includegraphics[width=\textwidth]{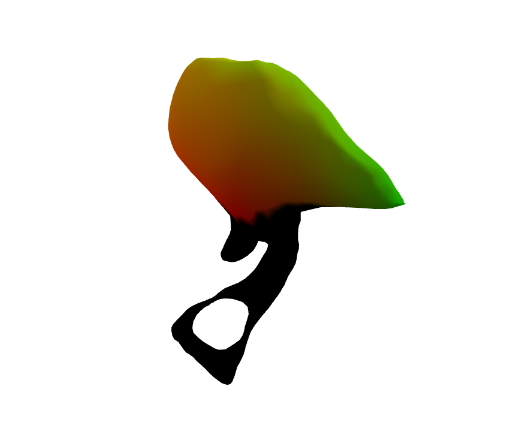}
    \end{subfigure}
    \hspace{.01\textwidth}
    \begin{subfigure}[b]{0.3\textwidth}
        \includegraphics[width=\textwidth]{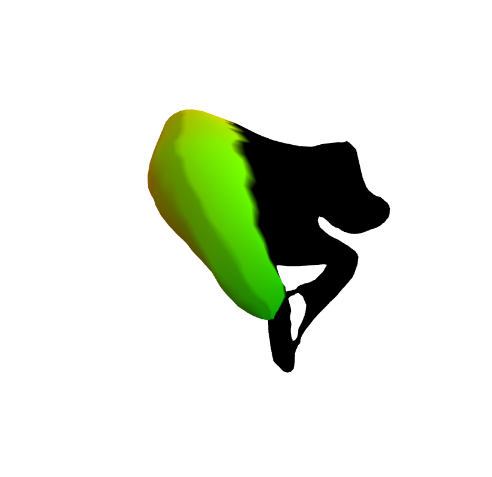}
    \end{subfigure}
    \hspace{.01\textwidth}
    \begin{subfigure}[b]{0.3\textwidth}
        \includegraphics[width=\textwidth]{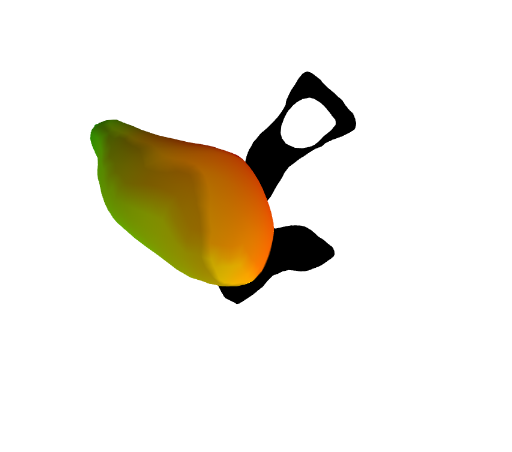}
    \end{subfigure}
    \caption{Generated coordinate map for surface mapping}
    \label{fig:coordinate_map}
\end{figure}

%% file: figures/video_frame.tex
\begin{figure}[H]
    \centering
    \begin{subfigure}[b]{0.23\textwidth}
        \includegraphics[width=\textwidth]{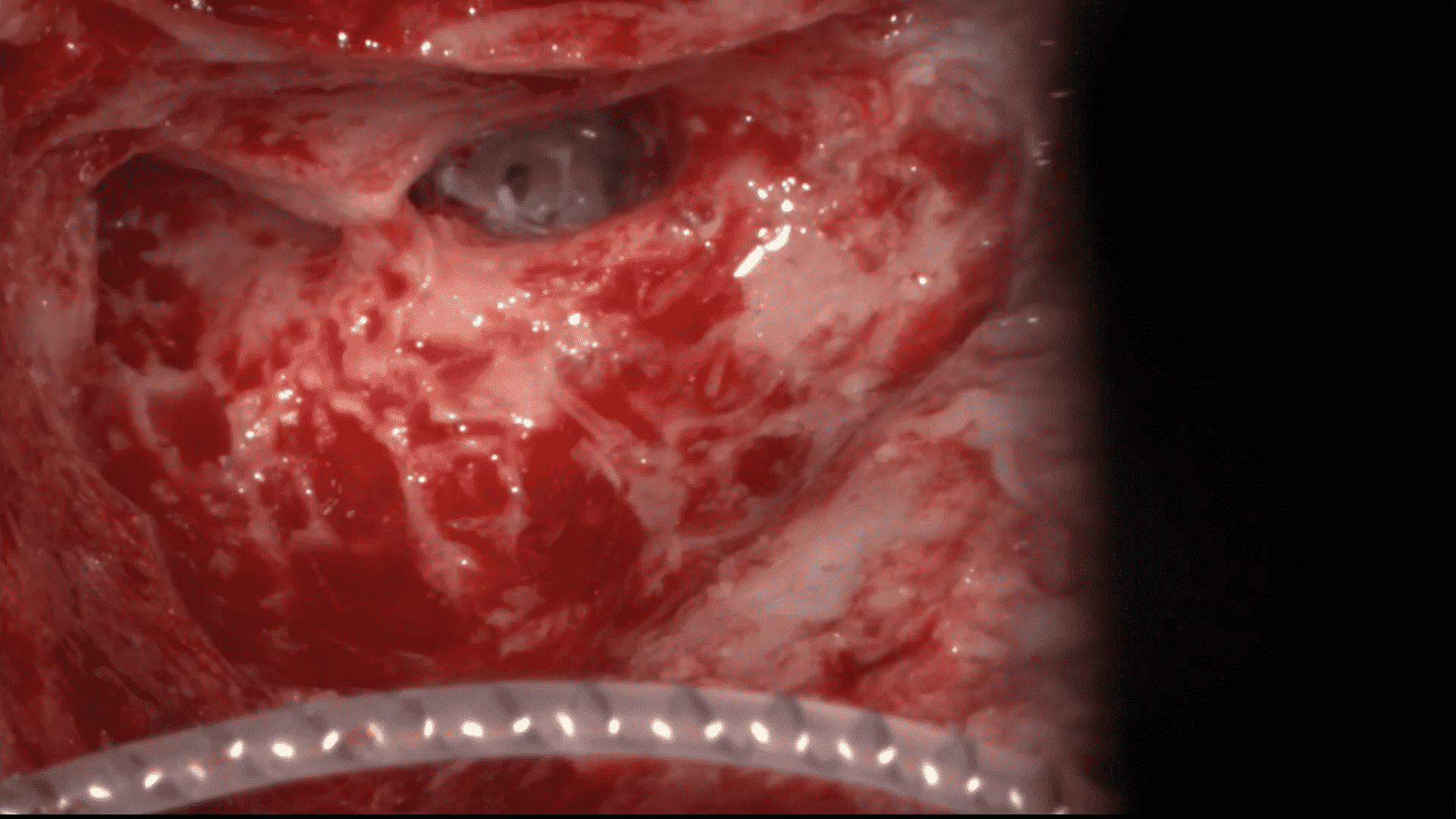}
        \caption{Original image}
        \label{subfig:o1}
    \end{subfigure}
    \hspace{.01\textwidth}  
    \begin{subfigure}[b]{0.23\textwidth}
        \includegraphics[width=\textwidth]{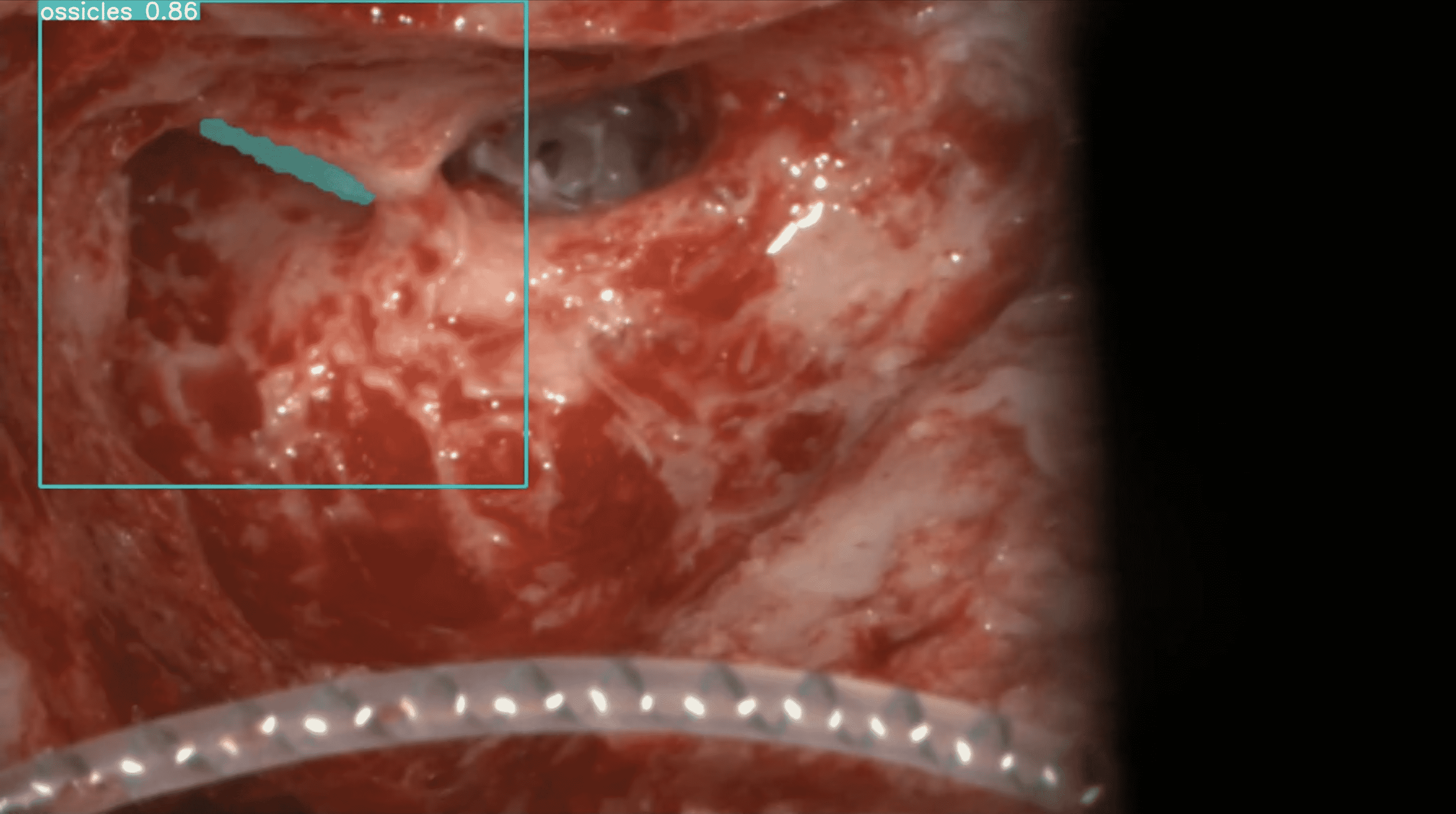}
        \caption{Bounding box and mask}
        \label{subfig:seg1}
    \end{subfigure}
    \hspace{.01\textwidth}
    \begin{subfigure}[b]{0.23\textwidth}
        \includegraphics[width=\textwidth]{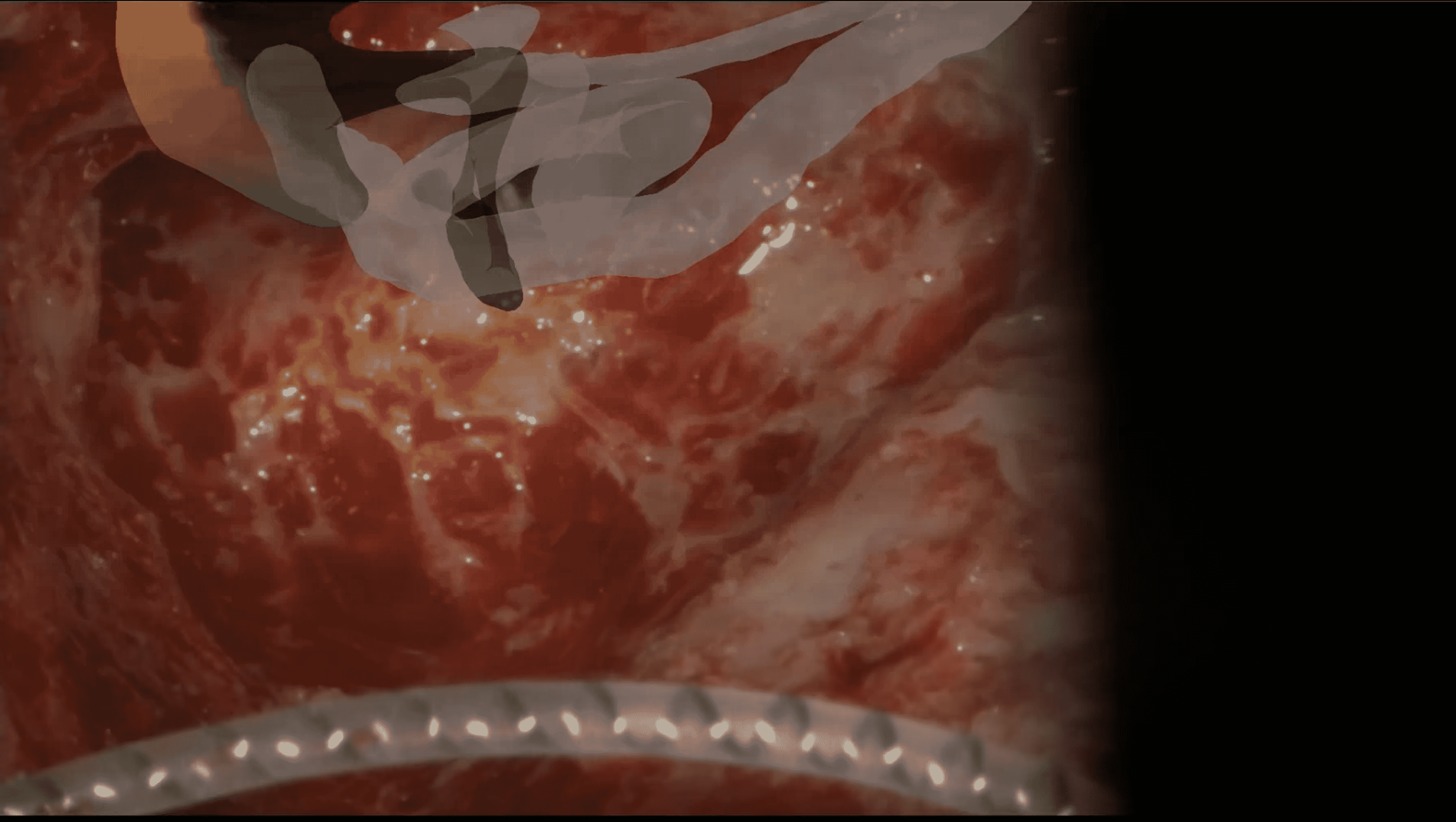}
        \caption{Manual registration}
        \label{subfig:register1}
    \end{subfigure}
    \hspace{.01\textwidth}
    \begin{subfigure}[b]{0.23\textwidth}
        \includegraphics[width=\textwidth]{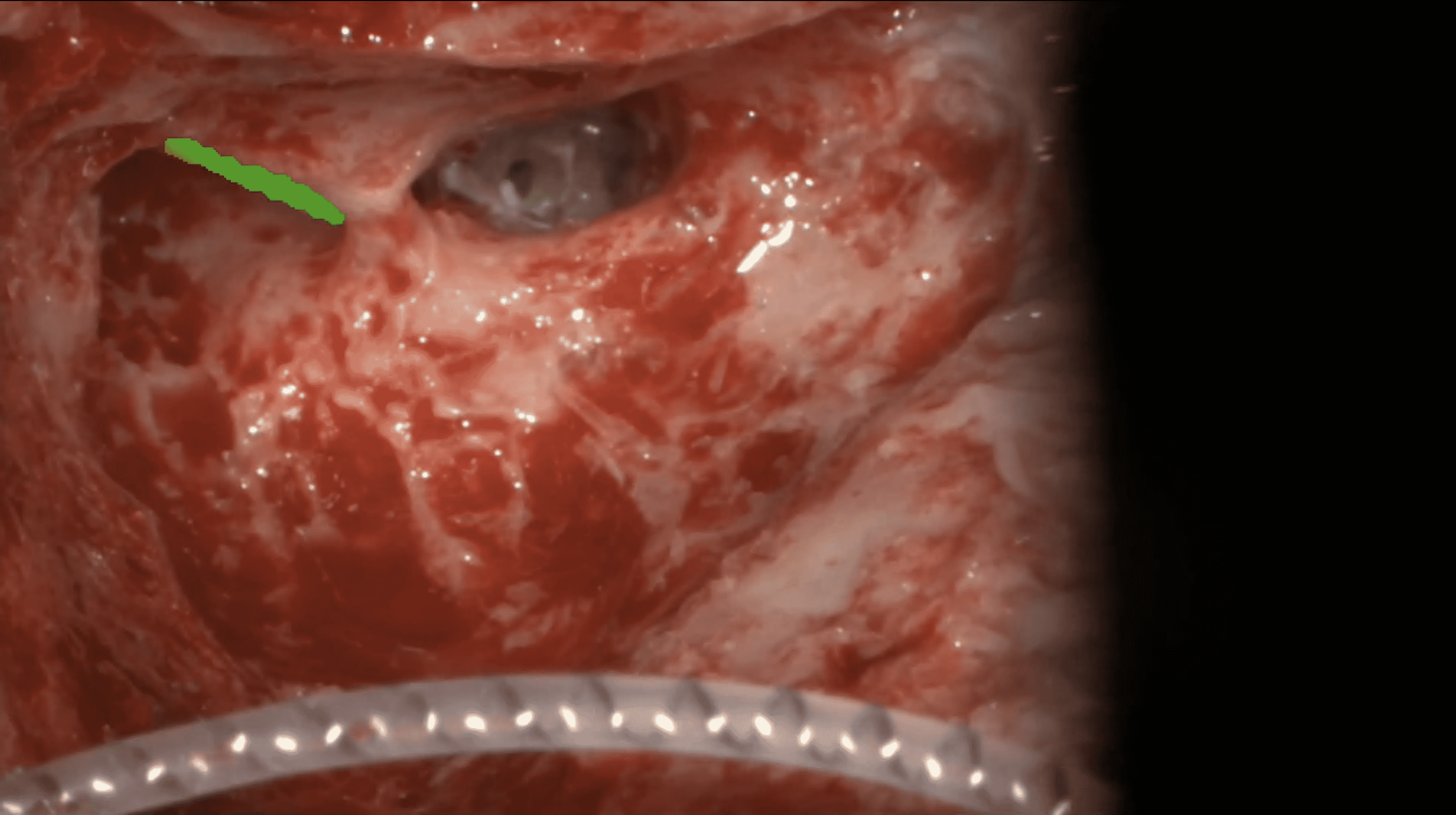}
        \caption{Visible surface on image}
        \label{subfig:visible1}
    \end{subfigure}
    \begin{subfigure}[b]{0.23\textwidth}
        \includegraphics[width=\textwidth]{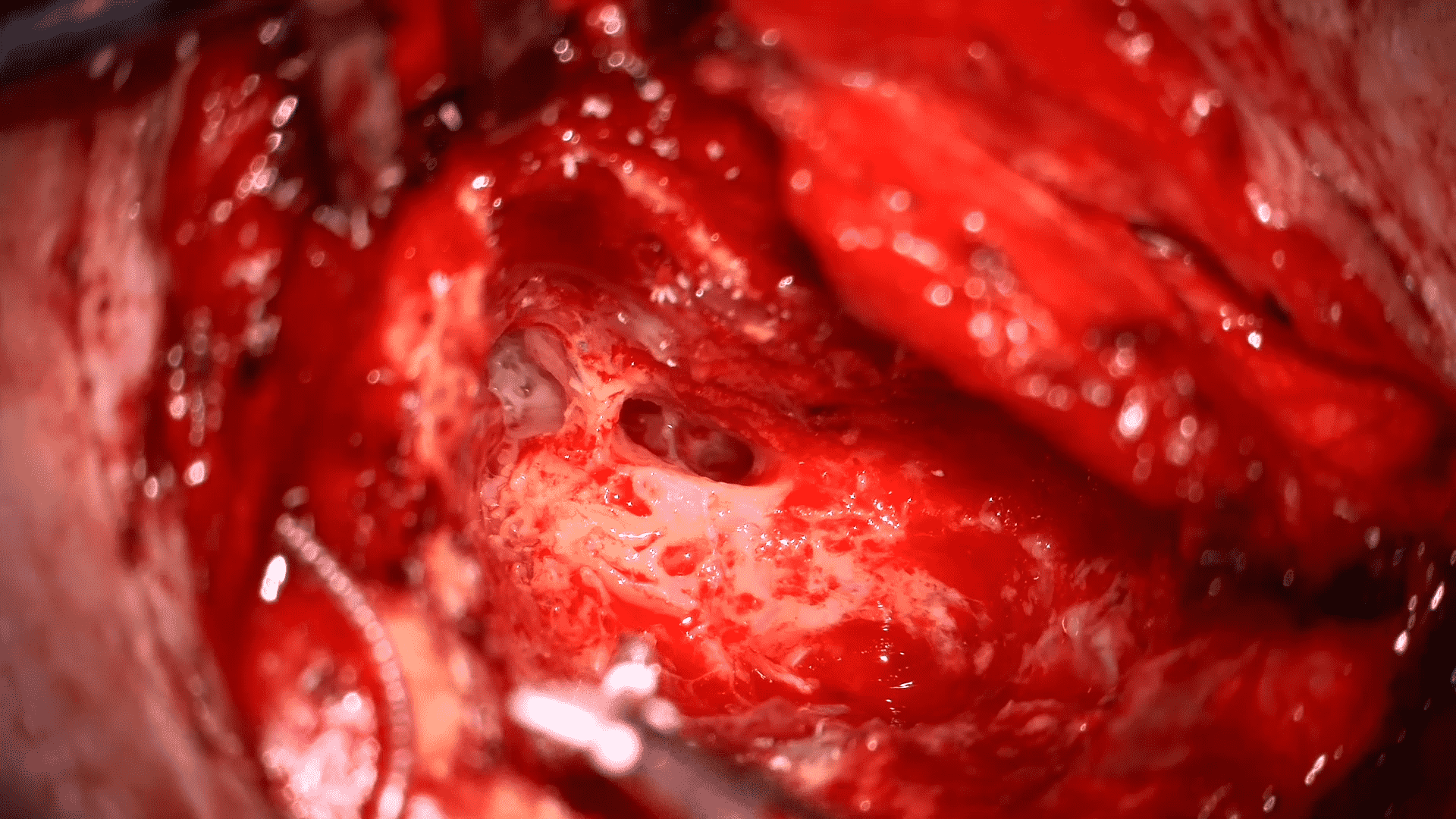}
        \caption{Original image}
        \label{subfig:o2}
    \end{subfigure}
    \hspace{.01\textwidth}  
    \begin{subfigure}[b]{0.23\textwidth}
        \includegraphics[width=\textwidth]{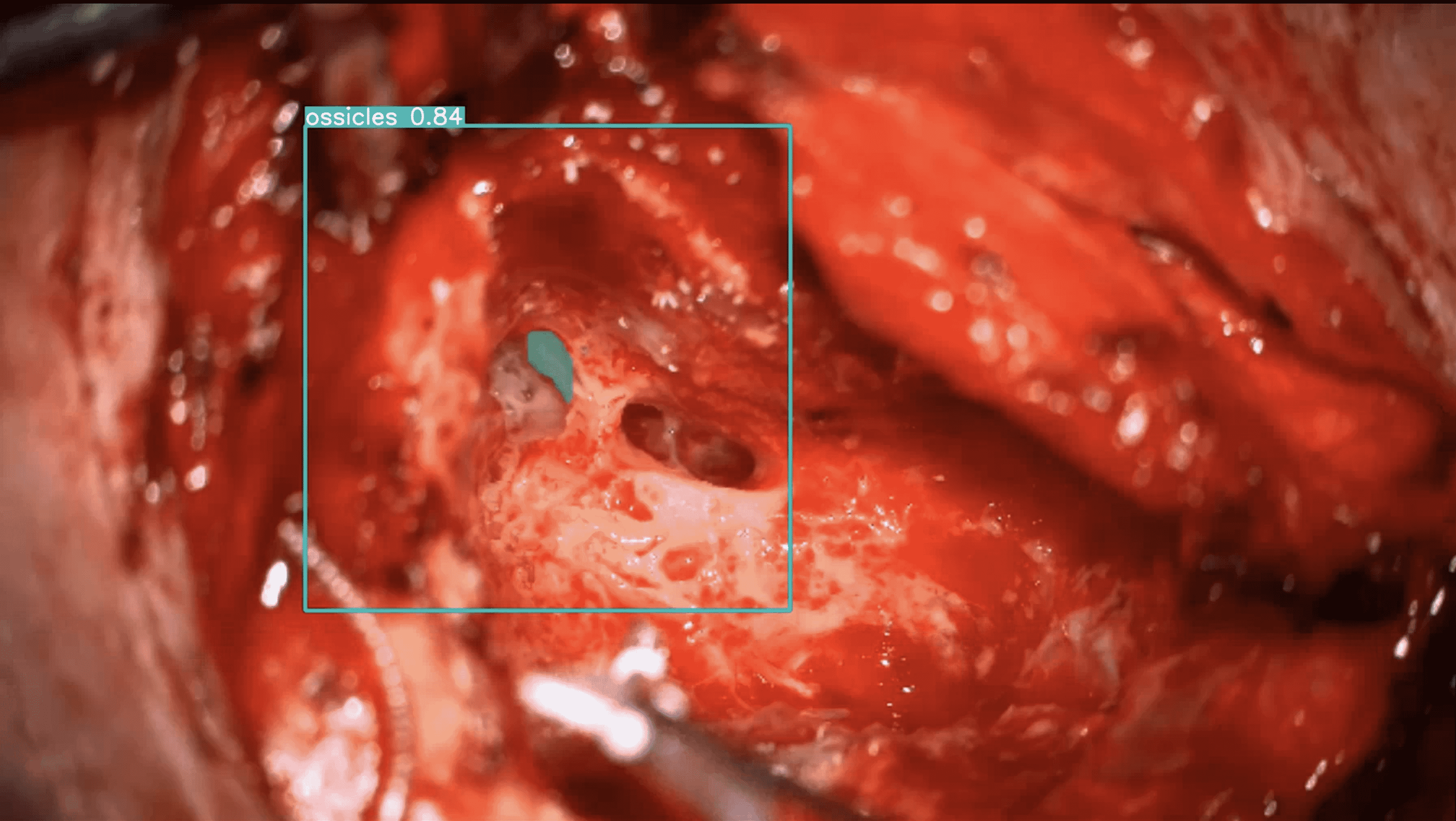}
        \caption{Bounding box and mask}
        \label{subfig:seg2}
    \end{subfigure}
    \hspace{.01\textwidth}
    \begin{subfigure}[b]{0.23\textwidth}
        \includegraphics[width=\textwidth]{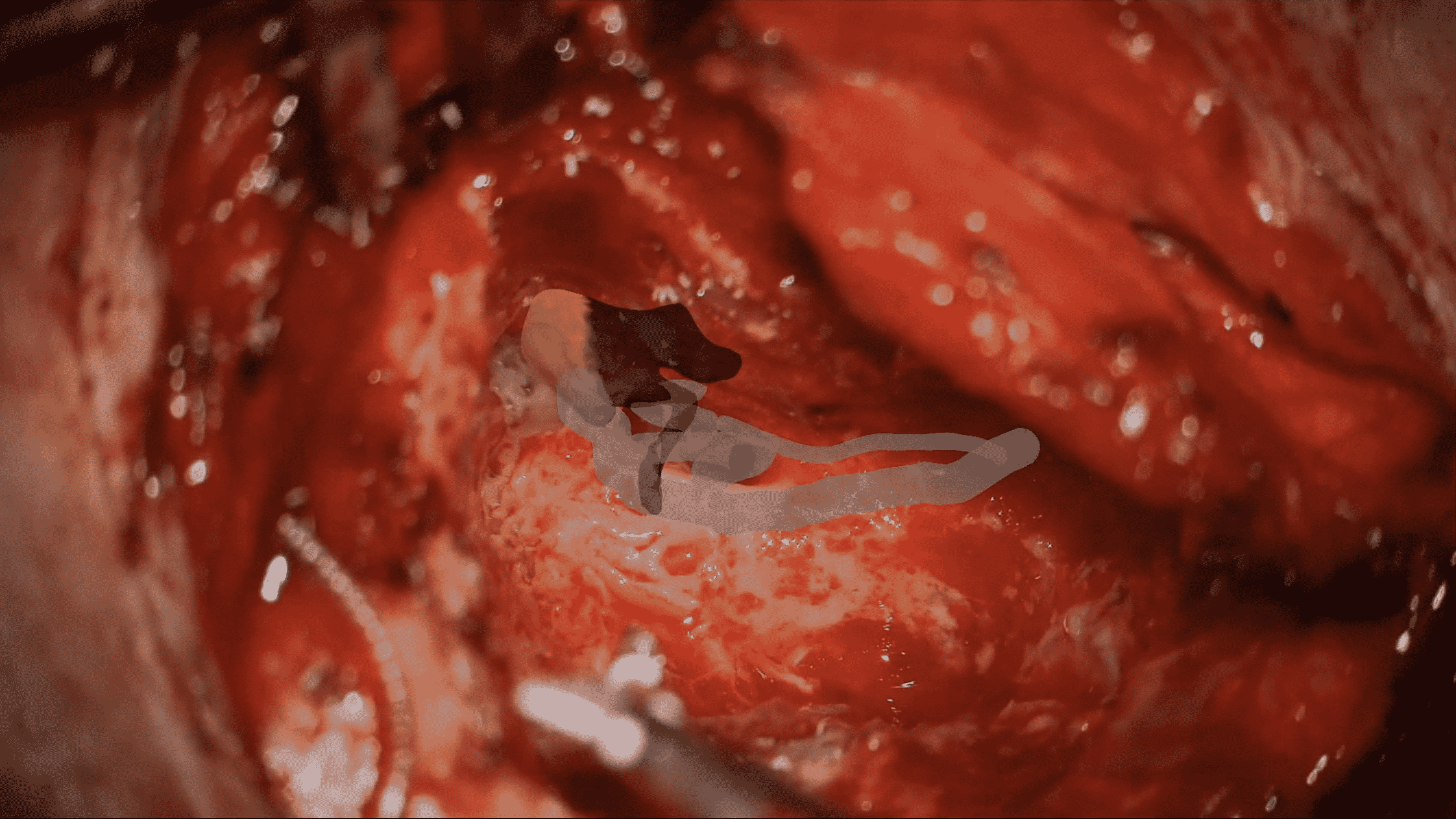}
        \caption{Manual registration}
        \label{subfig:register2}
    \end{subfigure}
    \hspace{.01\textwidth}
    \begin{subfigure}[b]{0.23\textwidth}
        \includegraphics[width=\textwidth]{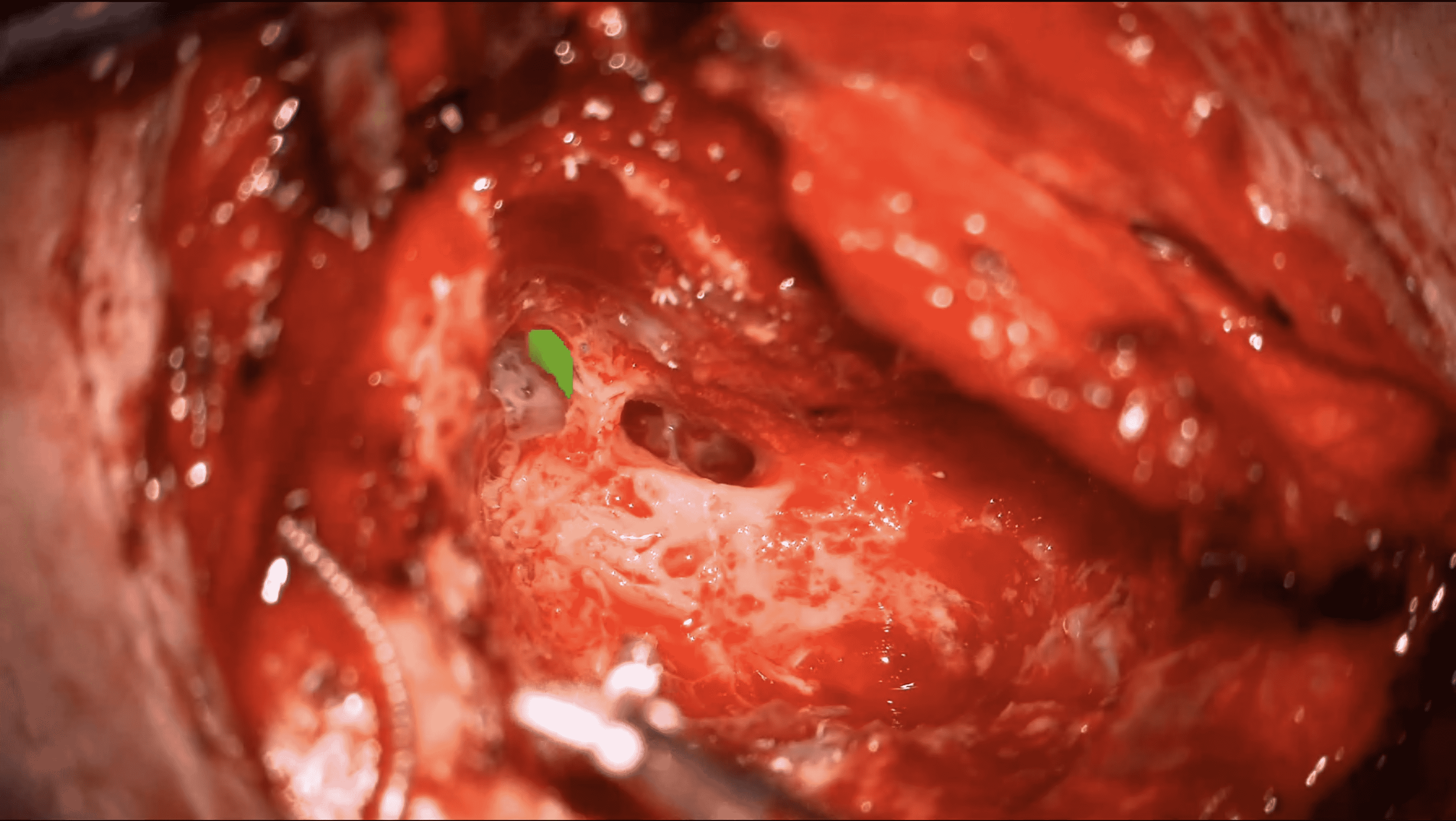}
        \caption{Visible surface on image}
        \label{subfig:visible2}
    \end{subfigure}
    \begin{subfigure}[b]{0.23\textwidth}
        \includegraphics[width=\textwidth]{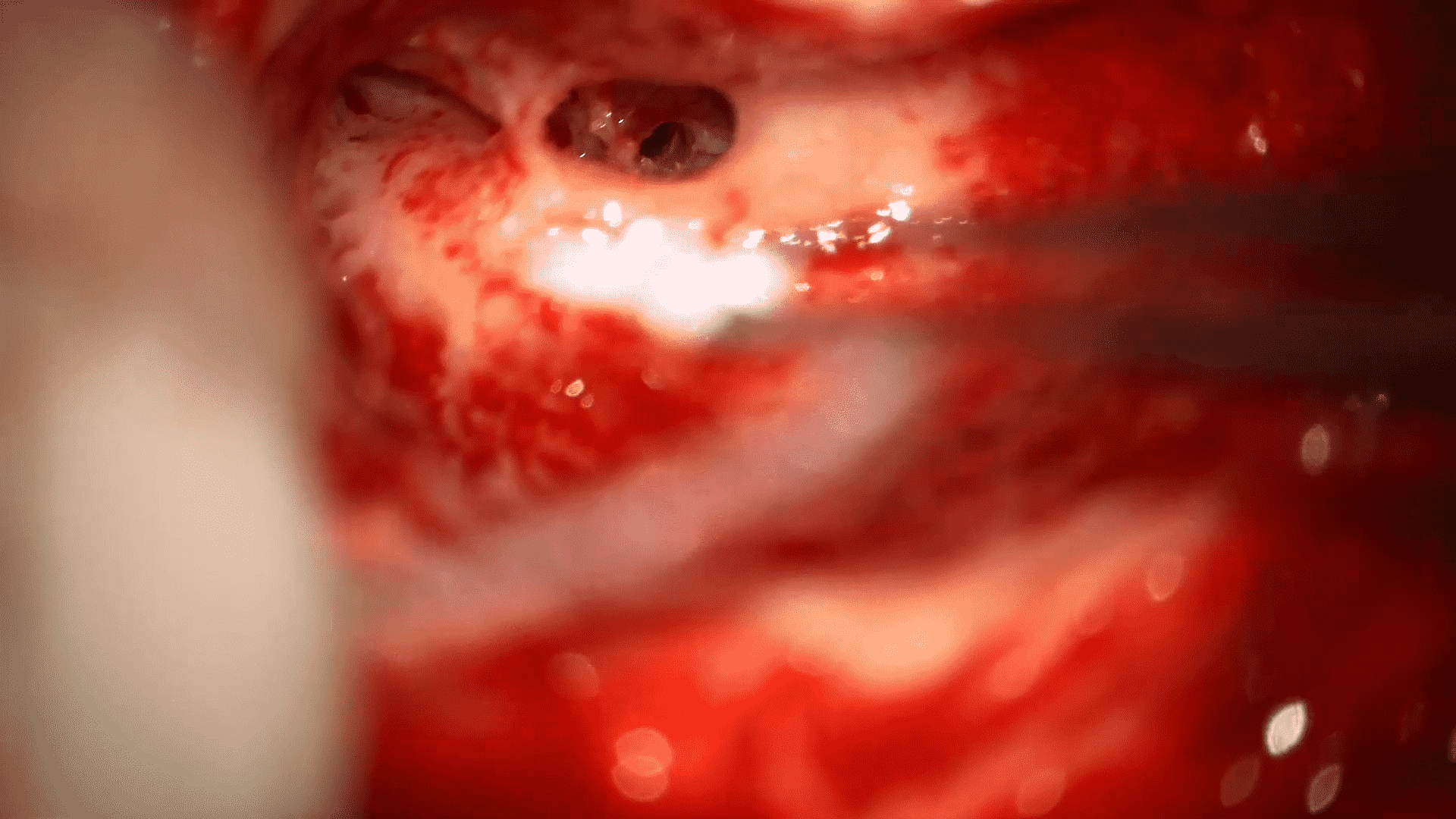}
        \caption{Original image}
        \label{subfig:o3}
    \end{subfigure}
    \hspace{.01\textwidth}  
    \begin{subfigure}[b]{0.23\textwidth}
        \includegraphics[width=\textwidth]{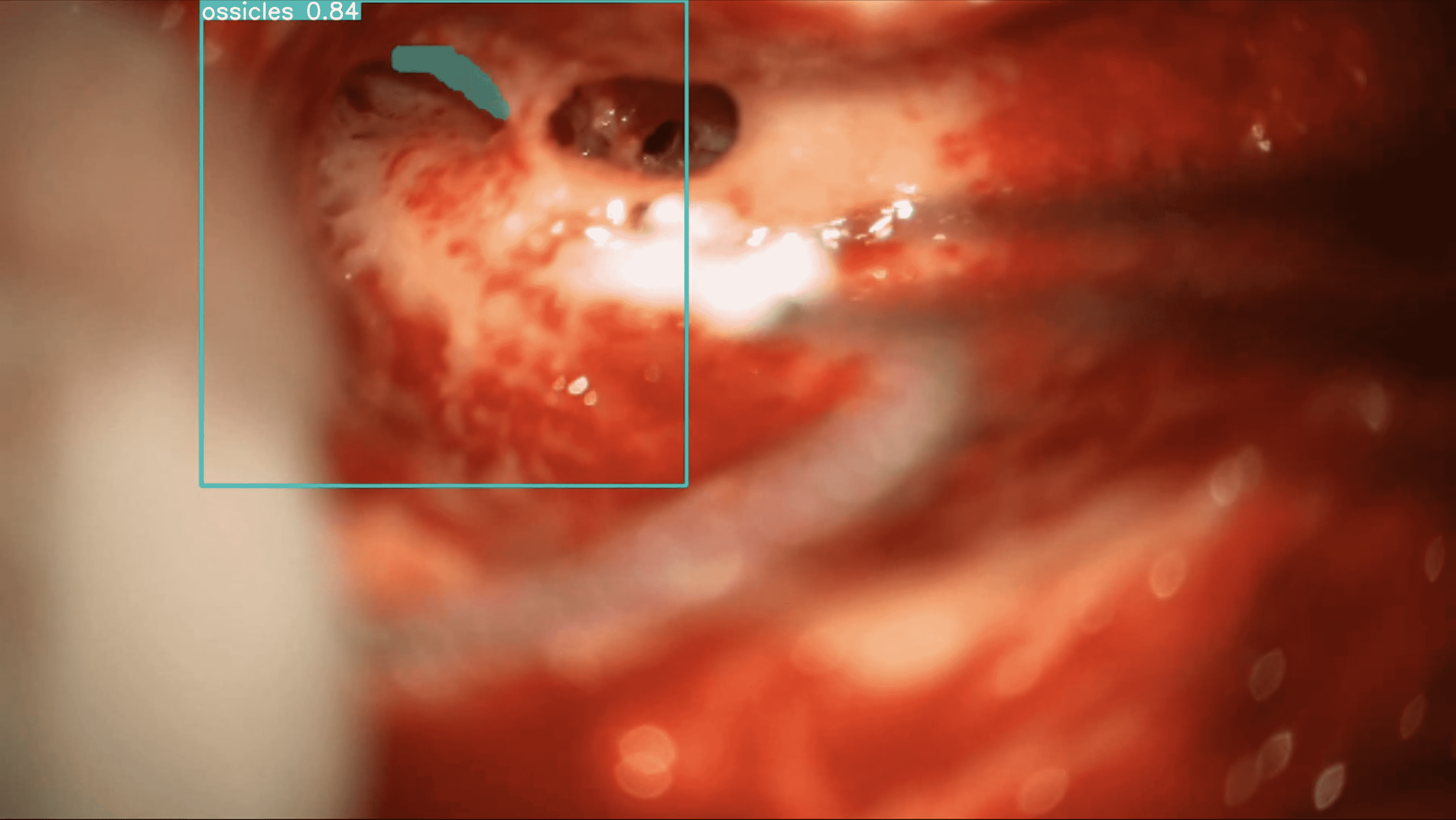}
        \caption{Bounding box and mask}
        \label{subfig:seg3}
    \end{subfigure}
    \hspace{.01\textwidth}
    \begin{subfigure}[b]{0.23\textwidth}
        \includegraphics[width=\textwidth]{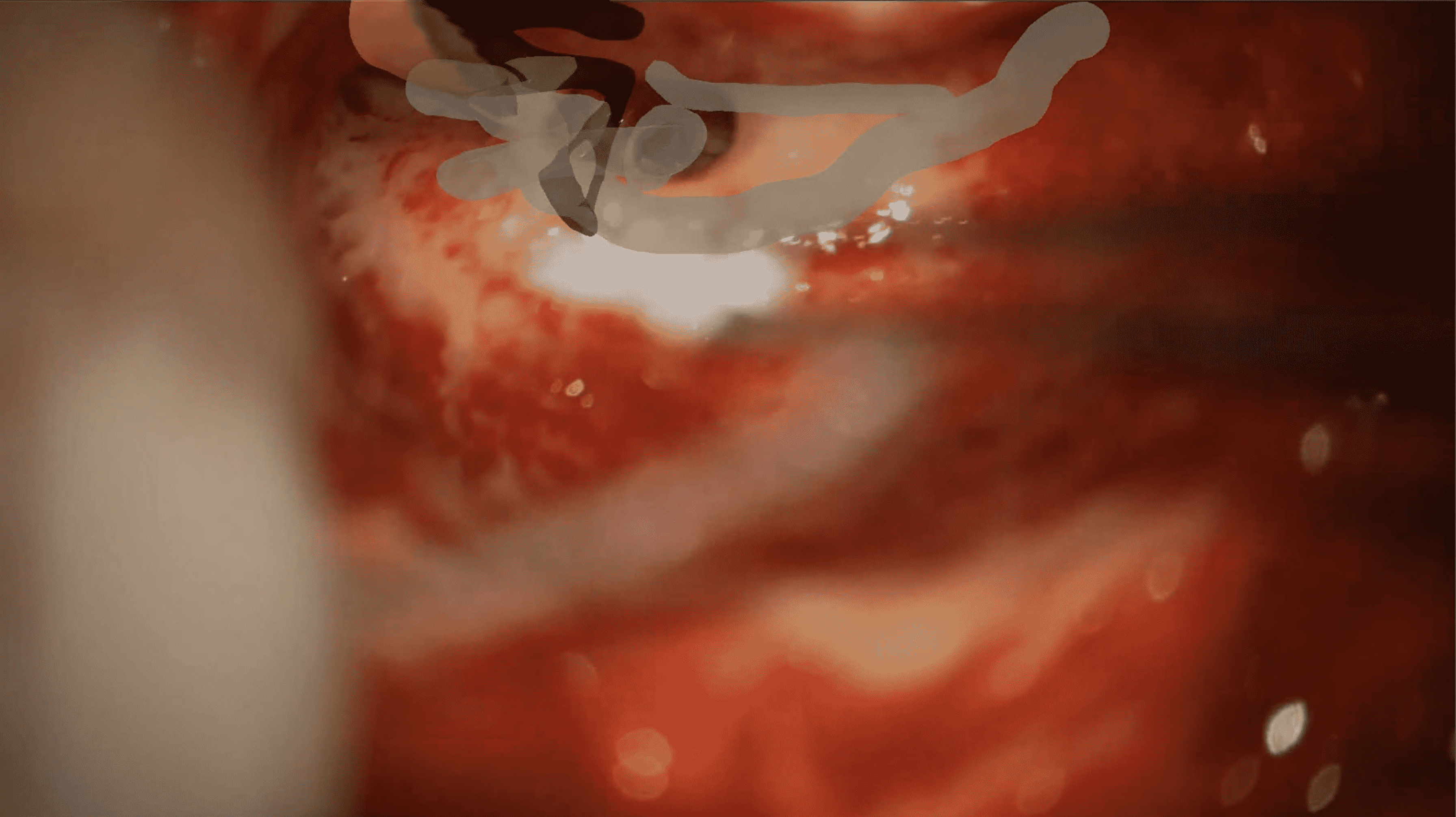}
        \caption{Manual registration}
        \label{subfig:register3}
    \end{subfigure}
    \hspace{.01\textwidth}
    \begin{subfigure}[b]{0.23\textwidth}
        \includegraphics[width=\textwidth]{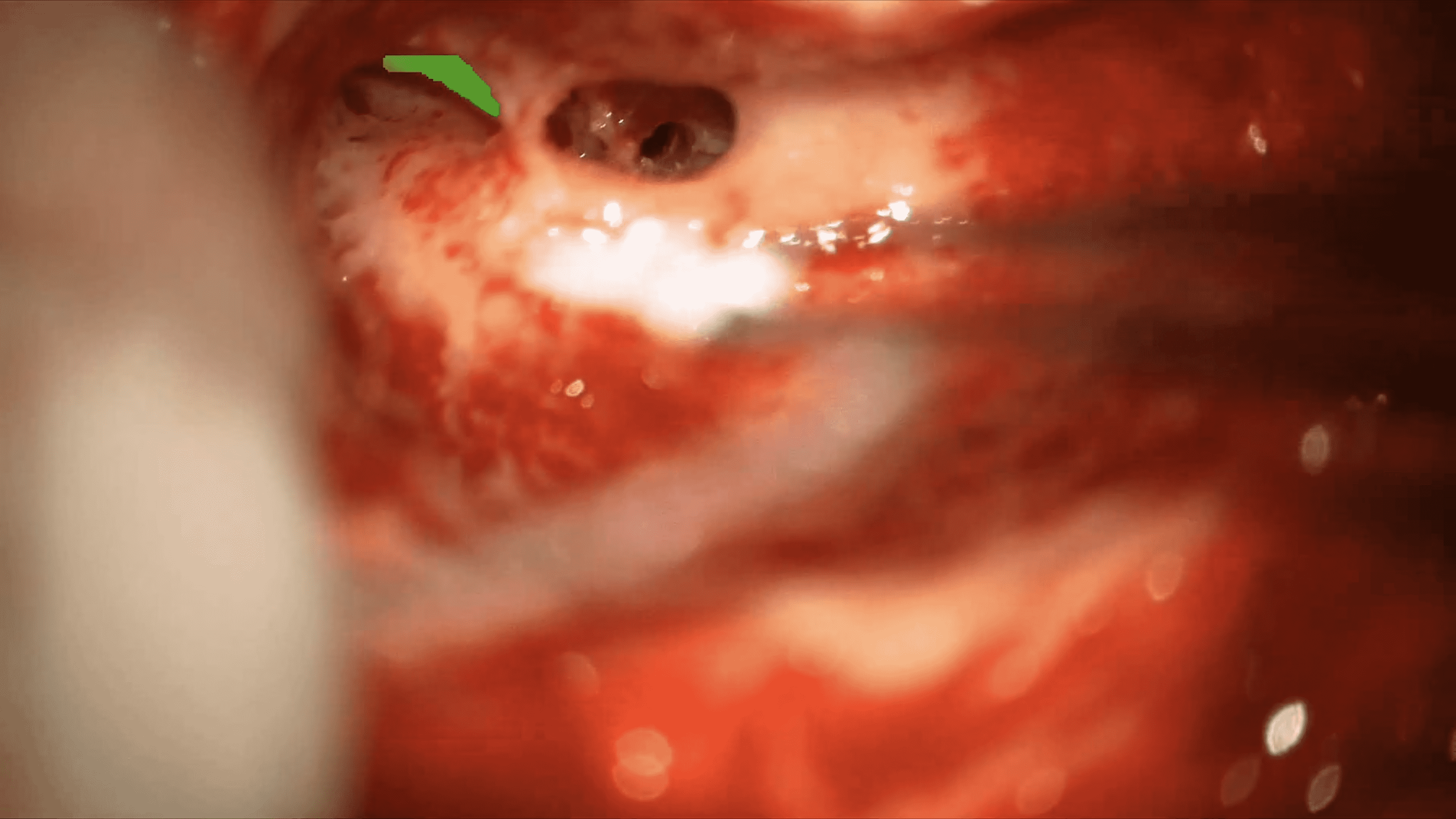}
        \caption{Visible surface on image}
        \label{subfig:visible3}
    \end{subfigure}
    \caption{Three samples random selected in data preparation process}
    \label{fig:video_frame}
\end{figure}

%% file: equations/loss.tex
\begin{align}
  Loss_{BCE} &= -\frac{1}{n} \sum_{i=1}^{n} \left( \rho_{i} \log(\sigma_{i}) + (1 - \rho_{i}) \log(1 - \sigma_{i}) \right)
  \label{eq:bce_loss}\\
  Loss_{MSE} &= \frac{1}{n} \sum_{i=1}^{n} (\rho_{i} - \sigma_{i})^{2} 
  \label{eq:mse_loss}\\
  Loss_{SSIM} &= [l(\rho,\sigma)]^\alpha \cdot [c(\rho,\sigma)]^\beta \cdot [s(\rho,\sigma)]^\gamma \label{eq:ssim_loss}
\end{align} The variable $\rho$ represents the ground truth coordinate map, whereas $\sigma$ corresponds to the coordinate map predicted by the network. In the SSIM loss function, $l(\rho,\sigma)$ is the luminance comparison function, $c(\rho,\sigma)$ is the contrast comparison function, and $s(\rho,\sigma)$ is the structure comparison function.


%% file: results.tex
\section{RESULTS}
\label{sec:results}
In Fig. \ref{fig:render}, we present the results of three different testing samples. These samples were excluded from the training process and demonstrate the performance of our model in dealing with unseen data. Each row corresponds to a unique test sample and is composed of four types of images. For each sample, Figs. \ref{subfig:gt_color_mask_1}, \ref{subfig:gt_color_mask_2}, \ref{subfig:gt_color_mask_3} include ground-truth color masks. These masks contain the ground-truth 2D-3D correspondence map for the target structure ossicles. Figs. \ref{subfig:color_mask_1}, \ref{subfig:color_mask_2}, \ref{subfig:color_mask_3} represent the predicted results of our proposed neural network. Figs. \ref{subfig:gt_color_mask_render_1}, \ref{subfig:gt_color_mask_render_2}, \ref{subfig:gt_color_mask_render_3} are renderings of the pose estimation. These were solved using the iterative PnP algorithm applied on ground-truth (GT) color masks. Finally, Figs. \ref{subfig:color_mask_render_1}, \ref{subfig:color_mask_render_2}, \ref{subfig:color_mask_render_3} were generated in the same way, but instead using the color masks predicted by the network. The last row suggests that the limited and small visible region negatively impacts performance compared to the previous two cases.
\input{figures/render}
\vspace{-.5cm}
A box plot is included in Fig. \ref{fig:box_plot}, and is analyzed on thirty color masks predicted by the network that are generated by three test samples. The rotation error is calculated using the angular distance in degrees, where $R1$ and $R2$ are two rotation matrices. Translation errors on the x and y axes are calculated using the Euclidean distance in millimeters. The z-axis translation error is calculated in percentage based on the estimated focal length. The demonstrated z-translation error results are relatively small compared with the focal length.
\input{figures/box_plot}

%% file: figures/render.tex
\setlength{\fboxsep}{0pt}
\begin{figure}[H]
    \centering
    \begin{subfigure}[b]{0.23\textwidth}
        \includegraphics[width=\textwidth]{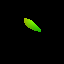}
        \caption{GT color mask}
        \label{subfig:gt_color_mask_1}
    \end{subfigure}
    \hspace{.008\textwidth}
    \begin{subfigure}[b]{0.23\textwidth}
        \includegraphics[width=\textwidth]{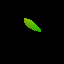}
        \caption{Predicted color mask}
        \label{subfig:color_mask_1}
    \end{subfigure}
    \hspace{.008\textwidth}
    \begin{subfigure}[b]{0.23\textwidth}
        \fbox{\includegraphics[width=\textwidth, height=3.92cm]{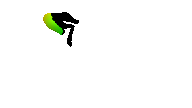}}
        \caption{GT pose render}
        \label{subfig:gt_color_mask_render_1}
    \end{subfigure}
    \hspace{.008\textwidth}
    \begin{subfigure}[b]{0.23\textwidth}
        \fbox{\includegraphics[width=\textwidth, height=3.92cm]{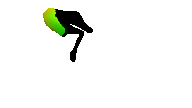}}
        \caption{Predicted pose render}
        \label{subfig:color_mask_render_1}
    \end{subfigure}

    \vspace{1em}
    
    \begin{subfigure}[b]{0.23\textwidth}
        \includegraphics[width=\textwidth]{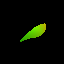}
        \caption{GT color mask}
        \label{subfig:gt_color_mask_2}
    \end{subfigure}
    \hspace{.008\textwidth}
    \begin{subfigure}[b]{0.23\textwidth}
        \fbox{\includegraphics[width=\textwidth]{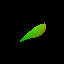}}
        \caption{Predicted color mask}
        \label{subfig:color_mask_2}
    \end{subfigure}
    \hspace{.008\textwidth}
    \begin{subfigure}[b]{0.23\textwidth}
        \fbox{\includegraphics[width=\textwidth, height=3.92cm]{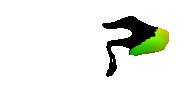}}
        \caption{GT pose render}
        \label{subfig:gt_color_mask_render_2}
    \end{subfigure}
    \hspace{.008\textwidth}
    \begin{subfigure}[b]{0.23\textwidth}
        \fbox{\includegraphics[width=\textwidth, height=3.92cm]{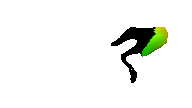}}
        \caption{Predicted pose render}
        \label{subfig:color_mask_render_2}
    \end{subfigure}
    
    \vspace{1em}

    \begin{subfigure}[b]{0.23\textwidth}
        \includegraphics[width=\textwidth]{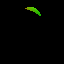}
        \caption{GT color mask}
        \label{subfig:gt_color_mask_3}
    \end{subfigure}
    \hspace{.008\textwidth}
    \begin{subfigure}[b]{0.23\textwidth}
        \includegraphics[width=\textwidth]{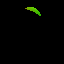}
        \caption{Predicted color mask}
        \label{subfig:color_mask_3}
    \end{subfigure}
    \hspace{.008\textwidth}
    \begin{subfigure}[b]{0.23\textwidth}
        \fbox{\includegraphics[width=\textwidth, height=3.92cm]{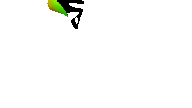}}
        \caption{GT pose render}
        \label{subfig:gt_color_mask_render_3}
    \end{subfigure}
    \hspace{.008\textwidth}
    \begin{subfigure}[b]{0.23\textwidth}
        \fbox{\includegraphics[width=\textwidth, height=3.92cm]{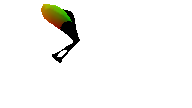}}
        \caption{Predicted pose render}
        \label{subfig:color_mask_render_3}
    \end{subfigure}
    \caption{Performance on three testing samples}
    \label{fig:render}
\end{figure}

%% file: figures/box_plot.tex
\begin{figure}[H]
    \centering
    \includegraphics[width=.95\textwidth]{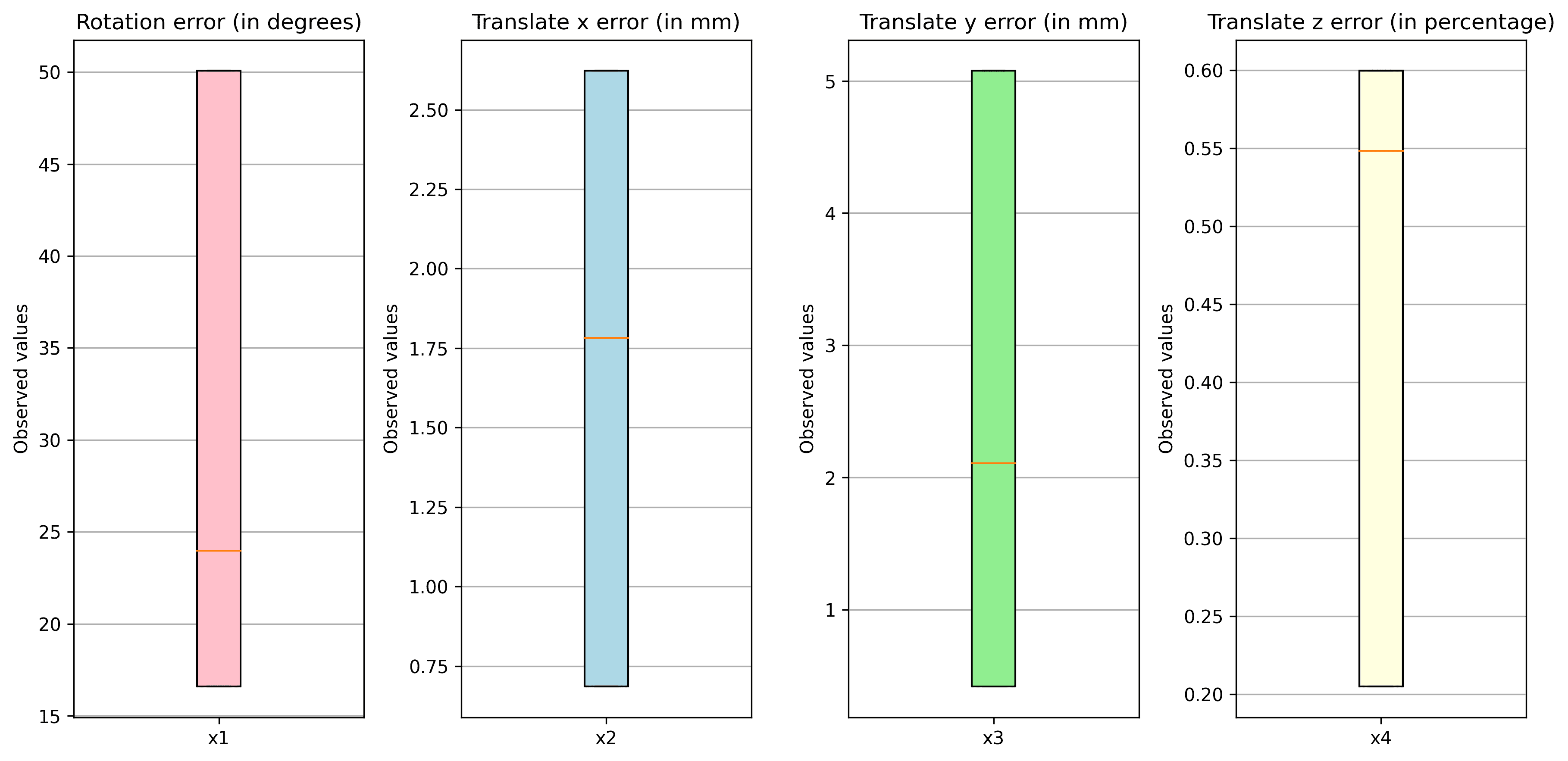}
    \caption{Reconstructed rotation error in degrees and translation error in mm.}
    \label{fig:box_plot}
\end{figure}

%% file: conclusion.tex
\section{CONCLUSION}
\label{sec:conclusion}
In this work, we introduce an effective two-step process designed to determine the pose of the incus in surgical scenes, even when it is severely occluded within cluttered surroundings. This technique represents an important advancement by providing an initial step in transitioning vital structural data and surgical plans from pre-operative CT scans to intra-operative surgical scenarios. Furthermore, our proposed method has the potential to be generalized, extending its applicability to various surgical procedures. From our demonstrated results, it is promising that the overall translation errors are low, the z translation error is essentially a scaling error that is very minimal as well compared with the estimated focal length. Rotation errors are high which is expected since the selected visible landmark is small and localized. In future work, we will continue to investigate other landmark structures that are not surgically altered and add them to the training process. We are confident that providing more information will reduce those errors.

%% file: acknowledgement.tex
\acknowledgments 
This work was supported in part by grants R01DC014037 and R01DC008408 from NIDCD. This work is solely the responsibility of the authors and does not necessarily reflect the views of this institute.

%% file: main.bbl
\begin{thebibliography}{1}

\bibitem{holden2013factors}
Holden, L.~K., Finley, C.~C., Firszt, J.~B., Holden, T.~A., Brenner, C., Potts, L.~G., Gotter, B.~D., Vanderhoof, S.~S., Mispagel, K., Heydebrand, G., and Skinner, M.~W., ``Factors affecting open-set word recognition in adults with cochlear implants,'' {\em Ear Hear}~{\bf 34},  342--360 (May-Jun 2013).

\bibitem{labadie2018preliminary}
Labadie, R. and Noble, J., ``Preliminary results with image-guided cochlear implant insertion techniques,'' {\em Otol Neurotol}~{\bf 39},  922--928 (Aug 2018).

\bibitem{vavra2017}
V{\'a}vra, P., Roman, J., Zon{\v{c}}a, P., Ihn{\'a}t, P., N{\v{e}}mec, M., Kumar, J., Habib, N., and El-Gendi, A., ``Recent development of augmented reality in surgery: A review,'' {\em J Healthc Eng}~{\bf 2017},  4574172 (2017).
\newblock Epub 2017 Aug 21.

\bibitem{dblp}
Zakharov, S., Shugurov, I., and Ilic, S., ``{DPOD:} dense 6d pose object detector in {RGB} images,'' {\em CoRR}~{\bf abs/1902.11020} (2019).

\bibitem{nocs}
Wang, H., Sridhar, S., Huang, J., Valentin, J., Song, S., and Guibas, L.~J., ``Normalized object coordinate space for category-level 6d object pose and size estimation,'' {\em CoRR}~{\bf abs/1901.02970} (2019).

\bibitem{zhang2023self}
Zhang, Y. and Noble, J.~H., ``Self-supervised registration and segmentation on ossicles with a single ground truth label,'' in [{\em Medical Imaging 2023: Image-Guided Procedures, Robotic Interventions, and Modeling}{\nolinebreak\hspace{0.1em}]},   {\bf 12466},  222--227, SPIE (2023).

\bibitem{Sethian}
{Sethian}, J.~A.,  [{\em {Level Set Methods and Fast Marching Methods}}{\nolinebreak\hspace{0.1em}]} (1999).

\bibitem{Jocher_YOLO_by_Ultralytics_2023}
Jocher, G., Chaurasia, A., and Qiu, J., ``{YOLO by Ultralytics},''  https://github.com/ultralytics/ultralytics (2023).

\bibitem{Zhang_Vision6D}
Zhang, Y., ``{Vision6D},''  https://github.com/ykzzyk/vision6D (2023).

\end{thebibliography}
